\documentclass[draftcls,onecolumn,11pt]{IEEEtran}
\usepackage{graphicx,epsfig,tabularx,hyperref,nccfoots}
\usepackage{xspace,bbm,epic,eepic,amssymb,amsmath,amsthm,bbm}
\usepackage{pdfpages,epstopdf,hyperref}
\def\acc#1{\left\{#1\right\}}              
\def\cro#1{\left[#1\right]}                
\def\bars#1{\left|#1\right|}

              \def\Card#1{{\mathrm{Card}}\cro{#1}}

\newsavebox{\fminibox}
\newlength{\fminilength}
\newenvironment{fminipage}[1][\linewidth]
  {\setlength{\fminilength}{#1}%-2\fboxsep-2\fboxrule}%
   \begin{lrbox}{\fminibox}\begin{minipage}{\fminilength}}
  {\end{minipage}\end{lrbox}\noindent\fbox{\usebox{\fminibox}}}

  \def\+{^\dagger}
\def\nequiv{\not\kern-.05em\equiv}
\def\egal{\kern-.5em=\kern-.5em}        % Moins d'espace autour de "="
\def\propt{\kern-.2em\propto\kern-.2em} % Idem
\def\argmin{\mathop{\mathrm{arg\,min}}} % car l'indice est reparti

\def\intdouble{\int\kern-0.3em\int}
\def\inttriple{\int\kern-0.3em\int\kern-0.3em\int}

\def\rond#1{\overset{\kern-0.33em~_\circ}{#1}}
\def\rondit[#1]#2{\overset{\kern#1~_\circ}{#2}}

\newtheorem{remark}{Remark}
\newtheorem{theorem}{Theorem}
\newtheorem{definition}{Definition}

\newtheorem{lemma}{Lemma}
\newtheorem{corollary}{Corollary}

\def\keywords{\vspace{-.3em}
    \if@twocolumn
      \small\it Keywords\/\bf---$\!$%
    \else
      \begin{center}\small\bf Keywords\end{center}\quotation\small
    \fi}
\def\endkeywords{\vspace{0.6em}\par\if@twocolumn\else\endquotation\fi
    \normalsize\rm}

\input{alphabet}
\def\babs{\begin{abstract}}             \def\eabs{\end{abstract}}
\def\barr{\begin{array}}                \def\earr{\end{array}}
\def\bcc{\begin{center}}                \def\ecc{\end{center}}

\def\bdes{\begin{description}}          \def\edes{\end{description}}
\def\bdoc{\begin{document}}             \def\edoc{\end{document}}
\def\ben{\begin{enumerate}}             \def\een{\end{enumerate}}
\def\beqn{\begin{eqnarray}}             \def\eeqn{\end{eqnarray}}
\def\beqnl#1{\beqn\label{#1}}           \def\eeqnl#1{\label{#1}\eeqn}
\def\beqnx{\begin{eqnarray*}}           \def\eeqnx{\end{eqnarray*}}
\def\bseqn{\begin{subeqnarray}}         \def\eseqn{\end{subeqnarray}}
\def\beq#1\eeq{\begin{equation}#1\end{equation}}
\def\bal#1\eal{\begin{align}#1\end{align}}
\def\balx#1\ealx{\begin{align*}#1\end{align*}}
\def\beqx{$$}                           \def\eeqx{$$}
\def\bfig{\protect\begin{figure}}       \def\efig{\protect\end{figure}}
\def\bfigx{\protect\begin{figure*}}     \def\efigx{\protect\end{figure*}}
\def\bfigt{\protect\begin{figurette}}   \def\efigt{\protect\end{figurette}}
\def\bfl{\begin{flushleft}}             \def\efl{\end{flushleft}}
\def\bfr{\begin{flushright}}            \def\efr{\end{flushright}}
\def\bit{\begin{itemize}}               \def\eit{\end{itemize}}
\def\bmi{\begin{minipage}}              \def\emi{\end{minipage}}
\def\bfmi{\begin{fminipage}}            \def\efmi{\end{fminipage}}
\def\bpic{\begin{picture}}              \def\epic{\end{picture}}
\def\bqu{\begin{quote}}                 \def\equ{\end{quote}}
\def\bqun{\begin{quotation}}            \def\equn{\end{quotation}}
\def\bsl{\begin{slide}}                 \def\esl{\end{slide}}
\def\btabb{\begin{tabbing}}             \def\etabb{\end{tabbing}}
\def\btabl{\begin{table}}               \def\etabl{\end{table}}
\def\btablx{\begin{table*}}             \def\etablx{\end{table*}}
\def\btabu{\begin{tabular}}             \def\etabu{\end{tabular}}
\def\btabx{\begin{tabular*}}            \def\etabx{\end{tabular*}}
\def\bbib{\begin{thebibliography}{999}} \def\ebib{\end{thebibliography}}
\def\bver{\begin{verbatim}}             \def\ever{\end{verbatim}}
\def\bca{\begin{cases}}                          \def\eca{\end{cases}}
\def\bfrm#1{\begin{frame}{#1}}                \def\efrm{\end{frame}}
\def\bblk#1{\begin{block}{#1}}                \def\eblk{\end{block}}
\input{abrege}

\begin{document}
\title{A sufficient condition on monotonic increase of the number of nonzero entry in the optimizer of $\ell$-1 norm penalized least-square problem}
\author{Junbo Duan\thanks{J. Duan and Y.-P. Wang are with the Department of Biomedical Engineering and Biostatistics, Tulane University, New Orleans, USA (e-mail: jduan@tulane.edu; wyp@tulane.edu).}, Charles Soussen\thanks{C. Soussen and D. Brie are with Centre de Recherche en Automatique de Nancy, Nancy University, Nancy, France (e-mail:
charles.soussen@cran.uhp-nancy.fr; david.brie@cran.uhp-nancy.fr).}, David Brie, J\'er\^ome Idier\thanks{J. Idier is with Institut de Recherche en Communication et Cybern¡äetique de Nantes, Nantes, France (e-mail:
jerome.idier@irccyn.ec-nantes.fr).} and Yu-Ping Wang}
\date{\today}
\maketitle

\begin{abstract}
The $\ell$-1 norm based optimization is widely used in signal processing, especially in recent compressed sensing theory. This paper studies the solution path of the $\ell$-1 norm penalized least-square problem, whose constrained form is known as Least Absolute Shrinkage and Selection Operator~(LASSO). A solution path is the set of all the optimizers with respect to the evolution of the hyperparameter (Lagrange multiplier). The study of the solution path is of great significance in viewing and understanding the profile of the tradeoff between the approximation and regularization terms. If the solution path of a given problem is known, it can help us to find the optimal hyperparameter under a given criterion such as the Akaike Information Criterion. In this paper we present a sufficient condition on $\ell$-1 norm penalized least-square problem. Under this sufficient condition, the number of nonzero entries in the optimizer or solution vector increases monotonically when the hyperparameter decreases. We also generalize the result to the often used total variation case, where the $\ell$-1 norm is taken over the first order derivative of the solution vector. We prove that the proposed condition has intrinsic connections with the condition given by Donoho \etal \cite{Donoho08} and the positive cone condition by Efron {\it el al} \cite{Efron04}. However, the proposed condition does not need to assume the sparsity level of the signal as required by Donoho \etal's condition, and is easier to verify than Efron \etal's positive cone condition when being used for practical applications.
\end{abstract}


%\tableofcontents

\section{Introduction}
The $\ell$-1 norm optimization problem received wildly focus in optimization and signal processing community in the last decade, especially in the context of compressed sensing,  because of its stable performance in sparse signal restoration \cite{Tropp04,Donoho06a}. The $\ell$-1 norm of a vector $\ub\in\Rbb^n$ is defined as:
\beqx
\|\ub\|_1=\sum_{i=1}^n|u_i|
\eeqx
where $u_i$ is the $i$-th entry of \ub and $|u_i|$ denotes the absolute value.

For a given observation $\yb\in\Rbb^m$, a common problem in compressed sensing theory is to estimate the sparse approximation $\yb\approx \Ab\ub$ in a given dictionary $\Ab\in\Rbb^{m\times n}$. The dictionary \Ab consists of the elementary signals we are interested in. Under Bayesian framework, when we assume Gaussian distribution on residual $\rb=\yb-\Ab\ub$ and Laplacian distribution on $\ub$, the above problem can be formulated as \cite{Nikolova07}:

\begin{equation}\label{homotopy}
\ub^*(\lambda)=\argmin_\ub \acc{E(\ub,\lambda)=\frac{1}{2}\|\yb-\Ab\ub\|^2+\lambda\|\ub\|_1}
\end{equation}
The constrained form reads
\begin{equation}\label{LASSO}
\argmin_\ub \frac{1}{2}\|\yb-\Ab\ub\|^2 \hbox{\hspace{1cm} subject to}~~\|\ub\|_1\leqslant \tau\tag{LASSO}
\end{equation}
which is well known in the literature as Least Absolute Shrinkage and Selection Operator~(LASSO). Because of the equivalence of the two forms as discussed in \cite{Vandenberg08}, all the results concerning the penalized form (\ie (\ref{homotopy})) in this paper can be applied straightforward to LASSO.

\begin{figure}\centering
\includegraphics[width=12cm]{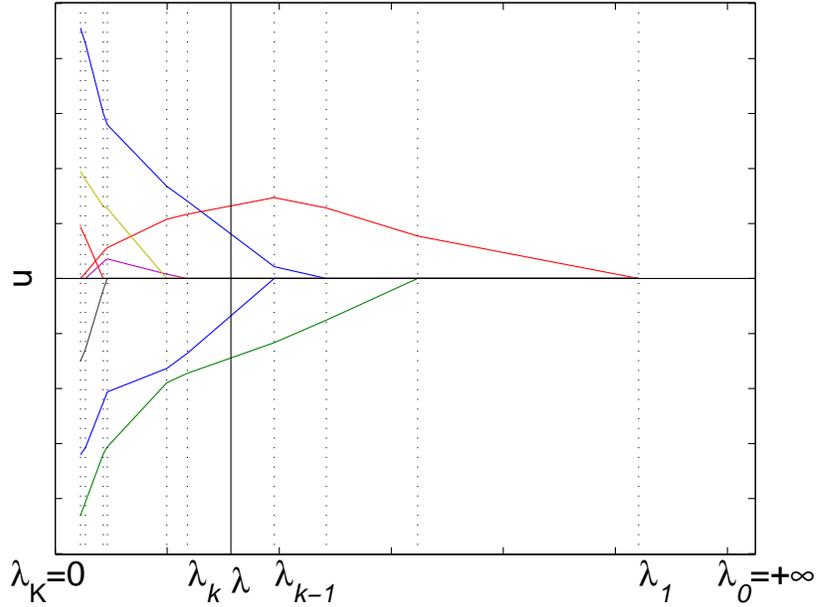}
\caption{A typical solution path of problem (\ref{homotopy}). Each colored curve corresponds to the evolution of an entry in \ub \wrt $\lambda$. To find the solution, the Homotopy and LARS algorithm usually start with $\lambda=+\infty$ and decrease the value step by step. Because it is piecewise linear at the interval $[\lambda_k, \lambda_{k-1}]$, the solution value at $\lambda$ can be evaluated from solutions at $\lambda_{k}$ and $\lambda_{k-1}$ by linear interpolation.}\label{fig_solutionpath}\end{figure}

The \textbf{solution path} of optimization problem (\ref{homotopy}) is defined as the set of all the optimizers \wrt the evolution of the hyperparameter: $\{\ub^*(\lambda)|\lambda\in (0, \infty)\}$. Fig. \ref{fig_solutionpath} shows a typical solution path. Each colored curve corresponds to an entry in \ub.

It is significant to find the solution path from both theoretical and application point of view. If the solution path is known, we can have the profile of the tradeoff between approximation term $\|\yb-\Ab\ub\|^2$ and regularization term $\|\ub\|_1$, which can help us to find the best hyperparameter under given criterion, such as L-curve \cite{Hansen92} or Akaike Information Criterion. For example, each $\lambda$ corresponds one data point $(\|\ub^*(\lambda)\|_1,\|\yb-\Ab\ub^*(\lambda)\|^2)$ at the 2D plane . All the data points form the Pareto frontier \cite{Vandenberg08}; and we can choose the data point having the largest curvature as the best tradeoff \cite{Hansen92}.

As a result of the discovery of the piecewise-linear-property of the solution path \cite{Tibshirani96}, algorithms like Homotopy \cite{Osborne00,Malioutov05} and Least Angle Regression~LARS~\cite{Efron04} were developed. The advantage of piecewise-linear-property is: If we have finite solutions $\{\ub^*(\lambda_k)|k=0,1,\ldots,K\}$, where $0=\lambda_K<\cdots<\lambda_1<\lambda_0=+\infty$ and $\ub^*(\lambda_k)(k=1,\ldots,K-1)$ is the solution at the boundary of two pieces, we can reconstruct the whole solution path for any $\lambda$. For any given hyperparameter $\lambda_k\leqslant\lambda<\lambda_{k-1}$, $\ub^*(\lambda)$ can be evaluated by linear interpolation:
\beqx
\ub^*(\lambda)=\ub^*(\lambda_k)+\frac{\lambda-\lambda_k}{\lambda_{k-1}-\lambda_k}\left(\ub^*(\lambda_{k-1})-\ub^*(\lambda_k)\right)
\eeqx

It is obvious that $\ub^*(+\infty)=\zerob$. As a result, Homotopy and LARS usually start with $\lambda=+\infty$ and decrease $\lambda$ step by step, as illustrated in Fig.\ref{fig_solutionpath}. In the iterations, critical value of $\lambda$, \ie $\lambda_k$ and the corresponding $\ub^*(\lambda_k)$ are calculated stepwisely. It is necessary to point out that, during the running of Homotopy algorithm, an active set $\Ic(\ub)=\{i|u_i\neq 0\}$ is maintained at each iteration, which updates the nonzero entries in \ub. If $u_i$ changes from zero to nonzero, we append \Ic with $i$; on the contrary, if $u_i$ changes from nonzero to zero, we remove $i$ from \Ic.

In previous work \cite{Donoho08}, Donoho \etal showed a condition on $\Ab$ and $\yb$ such that the number of element in set \Ic, \ie the cardinality $\Card{\Ic}$, \textbf{increases} monotonically when $\lambda$ \textbf{decreases}. This is known as \textbf{k-step solution property}, which is more strictly defined in Sec. \ref{subsec1}. So if $(\Ab,\yb)$ satisfies the condition yielding \textbf{k-step solution property}, \textbf{one only needs to appending the active set \Ic with a new entry}. Therefore, in each iteration of Homotopy algorithm, one only needs to check the change from zero to nonzero. Computation can thus be reduced. In other words, the Homotopy and LARS\footnote{Here we refer to the original version of LARS. The modified version of LARS which enable the removing of index from active set \Ic, is equivalent to Homotopy.} yield the same solution path. However, Donoho \etal's condition needs the knowledge of original signal \ub, \ie assuming the sparsity level, which is usually unknown in practical application. Therefore, in this paper we present a sufficient condition in which we do not assume the sparsity level of the signal.

This paper is organized as follows: In Sec.~\ref{sec2}, we present the sufficient condition on monotonic increase of $\Card{\Ic}$ when $\lambda$ decreases. In Sec.~\ref{sec3}, we discuss the connection between our proposed condition and other existing conditions. The total variation based approximation is often used in signal denoising~\cite{Chambolle97}, where the the $\ell$-1 norm is taken over the first order derivative of the solution vector. Therefore, in Sec.~\ref{sec4}, we extend the sufficient condition to the total variation case. We conclude the paper in Sec.~\ref{Conclu}.

\section{Sufficient condition}\label{sec2}
\begin{definition}\label{def1}
$\Hb\in\mathbb{R}^{n\times n}$ is called \textbf{(row) diagonally dominant~(DD)} if $h_{ii}\geqslant \sum_{j\neq i}|h_{ij}|,(i=1\ldots n)$; called \textbf{(row) irreducibly diagonally dominant~(IDD)} if at least one row meets $>$ instead of $\geqslant$; and \Hb is called \textbf{(row) strictly diagonally dominant~(SDD)} if all rows meet $>$ instead of $\geqslant$ \cite{Golub96}.
\end{definition}

\begin{definition}[Notations]
$\zerob_{k\times n}\in\mathbb{R}^{m\times n}$ is null matrix; $\Ib_{n}\in \mathbb{R}^{n\times n}$ is identity; $\Jb_{k\times n}=[\Ib_{k},\zerob_{k \times (n-k)}]\in\mathbb{R}^{k\times n}$; $\Pb$ is the square permutation matrix of size depending on the context; and $\Pb^T$ is the transpose of \Pb.
\end{definition}

\begin{lemma}[DD preservation property]\label{lemma1}
If full rank symmetric matrix $\Hb$ is DD, then $(\Jb_{k\times n}\Pb\Hb^{-1}\Pb^T\Jb_{k\times n}^T)^{-1}$ is also DD for any $\Pb$ and for all $k=1,\ldots, n$.
\end{lemma}

\begin{proof}
The Proofs can be found in \cite{LeiTGWLZ} and \cite{Carlson79}. However, we present a more comprehensible way of proof in Appendix.
\end{proof}

\begin{remark}
Lemma \ref{lemma1} indicates, for an DD matrix \Hb, if we invert it, extract the principal minor of any size $k\times k$, then the inverse of this principal minor is also DD.
\end{remark}

Based on Lemma \ref{lemma1}, we give our main result
\begin{theorem}\label{thm1}
For full rank matrix $\Ab\in\mathbb{R}^{m\times n}(m\geqslant n)$, in optimization problem (\ref{homotopy}), if $(\Ab^T\Ab)^{-1}$ is DD, $\Card{\Ic(\ub^*(\lambda))}$ \textbf{increases} monotonically when $\lambda$ \textbf{decreases}.
\end{theorem}

\begin{proof}
The differential of $E(\ub,\lambda)$ is:
\begin{equation}
\nonumber\partial E(\ub,\lambda)=\Ab^T(\Ab\ub(\lambda)-\yb)+\lambda \sb(\lambda)
\end{equation}
here $\sb$ is the subdifferential of $\|\ub\|_1$ \cite{Rockafellar70}, which is defined as:
\beq\label{def_subdif}
\sb=\partial\|\ub\|_1=\left\{
                                 \begin{array}{ll}
                                   s_i=1, & \hbox{if $u_i>0$;} \\
                                   s_i=-1, & \hbox{if $u_i<0$;} \\
                                   s_i\in[-1,1], & \hbox{otherwise.}
                                 \end{array}
                               \right.
\eeq
A necessary condition to the optimization problem (\ref{homotopy}) is to have $\zerob\in \partial E(\ub,\lambda)$; therefore, we have the following system:
\begin{equation}\label{sys}
\Ab^T\Ab\ub^*(\lambda)+\lambda\sb^*(\lambda)=\Ab^T\yb
\end{equation}
Because $\ub^*(\lambda)$ is piecewise linear \cite{Efron04}, for each piece $[\lambda_{k},\lambda_{k-1})$, $\sb^*(\lambda)$ is constant. Thus we can find a permutation $\Pb$ locally such that the nonzero entries and zero entries in $\ub^*$ are rearranged to be $\ub^*_{on}(\neq \zerob)$ and $\ub^*_{off}(=\zerob)$ respectively. In the following, we omit the dependency of $\lambda$ for the sake of brevity.
\begin{eqnarray}
\ub^*&=&\Pb^T\left[
\begin{array}{c}
\ub^*_{on}\\
\ub^*_{off}\end{array}
\right]\label{defu}\\
\sb^*&=&\Pb^T\left[
\begin{array}{c}
\sb^*_{on}\\
\sb^*_{off} \end{array}
\right]\label{defs}\\
\Ab^T\yb&=&\Pb^T\left[
\begin{array}{c}
\yb_{on}\\
\yb_{off} \end{array}
\right]\label{defAy}
\end{eqnarray}
By substituting (\ref{defu}), (\ref{defs}) and (\ref{defAy}) into (\ref{sys}), and left multiplying \Pb, since $\Pb^T=\Pb^{-1}$, we have
\begin{equation}
\Pb\Ab^T\Ab\Pb^T\left[\begin{array}{c}\ub^*_{on}\\\zerob\end{array}\right]
+\lambda\left[\begin{array}{c}\sb^*_{on}\\\sb^*_{off}\end{array}\right]
=\left[\begin{array}{c}\yb_{on}\\\yb_{off}\end{array}\right]
\end{equation}
which can be rewritten as:
\begin{equation}
\nonumber\left[\begin{array}{cc}\Psib&\Upsilonb\\\Upsilonb^T&\Phib\end{array}\right]\left[\begin{array}{c}\ub_{on}^*\\\zerob\end{array}\right]
+\lambda\left[\begin{array}{c}\sb_{on}^*\\\sb_{off}^*\end{array}\right]
=\left[\begin{array}{c}\yb_{on}\\\yb_{off}\end{array}\right]
\end{equation}
or
\beqn\label{sys1}
\Psib\ub_{on}^*+\lambda\sb_{on}^*&=&\yb_{on}\\
\nonumber\Upsilonb^T\ub_{on}^*+\lambda\sb_{off}^*&=&\yb_{off}
\eeqn
where $\Psib=\Jb_{k\times n}\Pb\Ab^T\Ab\Pb^T\Jb_{k\times n}^T$ and $k$ is the length of $\ub_{on}^*$. Under the condition that $(\Ab^T\Ab)^{-1}$ is DD, from Lemma \ref{lemma1}, $\Rb=\Psib^{-1}$ is DD. From (\ref{sys1})
\begin{equation}
\frac{d\ub_{on}^*}{d\lambda}=-\Rb\sb_{on}^*
\end{equation}
%Because $\Rb_k$ is assumed IDD, $r_{ii}\geqslant \sum_{j\neq i}|r_{ij}|,(i=1,\ldots,k)$
For the $i$-th entry of $\ub^*_{on}$, \ie $u_{on,i}^*(i=1,\ldots,k)$
\begin{equation}
\nonumber\frac{du^*_{on,i}}{d\lambda}=-\sum_{j=1}^kr_{ij}s^*_{on,j}=-r_{ii}s^*_{on,i}-\sum_{j\neq i}r_{ij}s^*_{on,j}
\end{equation}
because $s^*_{on,j}\in[-1,1]$, \\
(1) If $u_{on,i}^*>0$, from (\ref{def_subdif}) $s_{on,i}^*=1$\\
\begin{equation}
\nonumber\frac{du_{on,i}^*}{d\lambda}=-r_{ii}-\sum_{j\neq i}r_{ij}s^*_{on,j}\stackrel{\mathrm{DD}}{\leqslant}-\sum_{j\neq i}(|r_{ij}|+r_{ij}s^*_{on,j})\leqslant0
\end{equation}
(2) If $u^*_{on,i}<0$, from (\ref{def_subdif}) $s^*_{on,i}=-1$\\
\begin{equation}
\nonumber\frac{du_{on,i}^*}{d\lambda}=r_{ii}-\sum_{j\neq i}r_{ij}s^*_{on,j}\stackrel{\mathrm{DD}}{\geqslant}\sum_{j\neq i}(|r_{ij}|-r_{ij}s^*_{on,j})\geqslant0
\end{equation}
From above two cases, we can see that $|u^*_{on,i}(\lambda)|$ decreases monotonically when $\lambda$ increases in piece $[\lambda_{k},\lambda_{k-1})$, while $|u^*_{off,i}(\lambda)|$ is equal to zero.

Because $u_i^*(\lambda)$ is continuous for $\lambda>0$ \cite{Osborne00}, it is straightforward to extend the result to all $\lambda$: When $\lambda$ increases, the absolute value of nonzero entries in $\ub^*(\lambda)$ decrease until to 0, while zero entries remain 0. Therefore, $\Card{\Ic(\ub^*(\lambda))}$ decreases monotonically when $\lambda$ increases. In other words, $\Card{\Ic(\ub^*(\lambda))}$ increases monotonically when $\lambda$ decreases.
\end{proof}

There exist many matrices satisfying the sufficient condition. Obvious examples are the orthogonal dictionaries like Fourier basis or Hadamard basis. By Monte Carlo simulation, we also study the probability of random matrices satisfying the sufficient condition. For each given configuration $(m,n)$ and distribution \Pc, 1000 trials $\Ab\in\Rbb^{m\times n}$ are generated, whose entries obey \iid~\Pc. \Pc is chosen as: normal distribution, uniform distribution within interval $[0,1]$ and Bernoulli distribution with parameter $p=0.1,0.5$ (the probability for 1 is $p$, for 0 is $1-p$). The frequency of $(\Ab^T\Ab)^{-1}$ being DD is shown in Fig. \ref{fig1}. From the simulation results, we found that random matrices satisfy the sufficient condition when $m\gg n$.

In compressed sensing~(CS)~\cite{Donoho06,Candes08}, random matrix is frequently utilized to project a high dimension sparse signal into a low dimension space. If the correlation between the columns in the random matrix \Ab is low enough, and the original signal is also sparse enough, the original signal can be recovered from its observation \via $\ell$-1 optimization or other methods. In the next section, we show the intrinsic connection between our result and those derived by Donoho \etal\cite{Donoho08} and and Efron \etal\cite{Efron04} in CS theory.

\bfig
\centering
\btabu{cc}
\includegraphics[width=7cm]{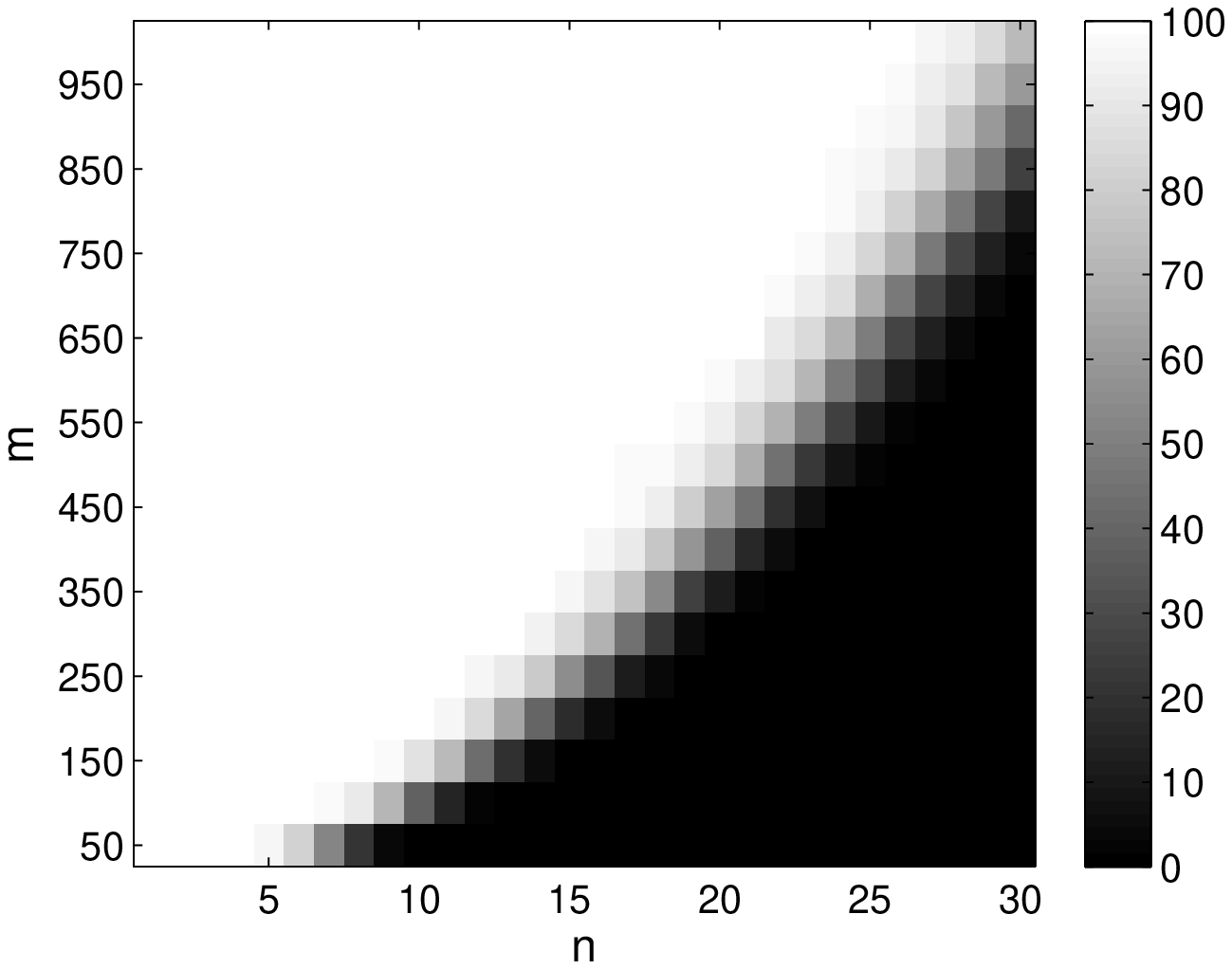}&\includegraphics[width=7cm]{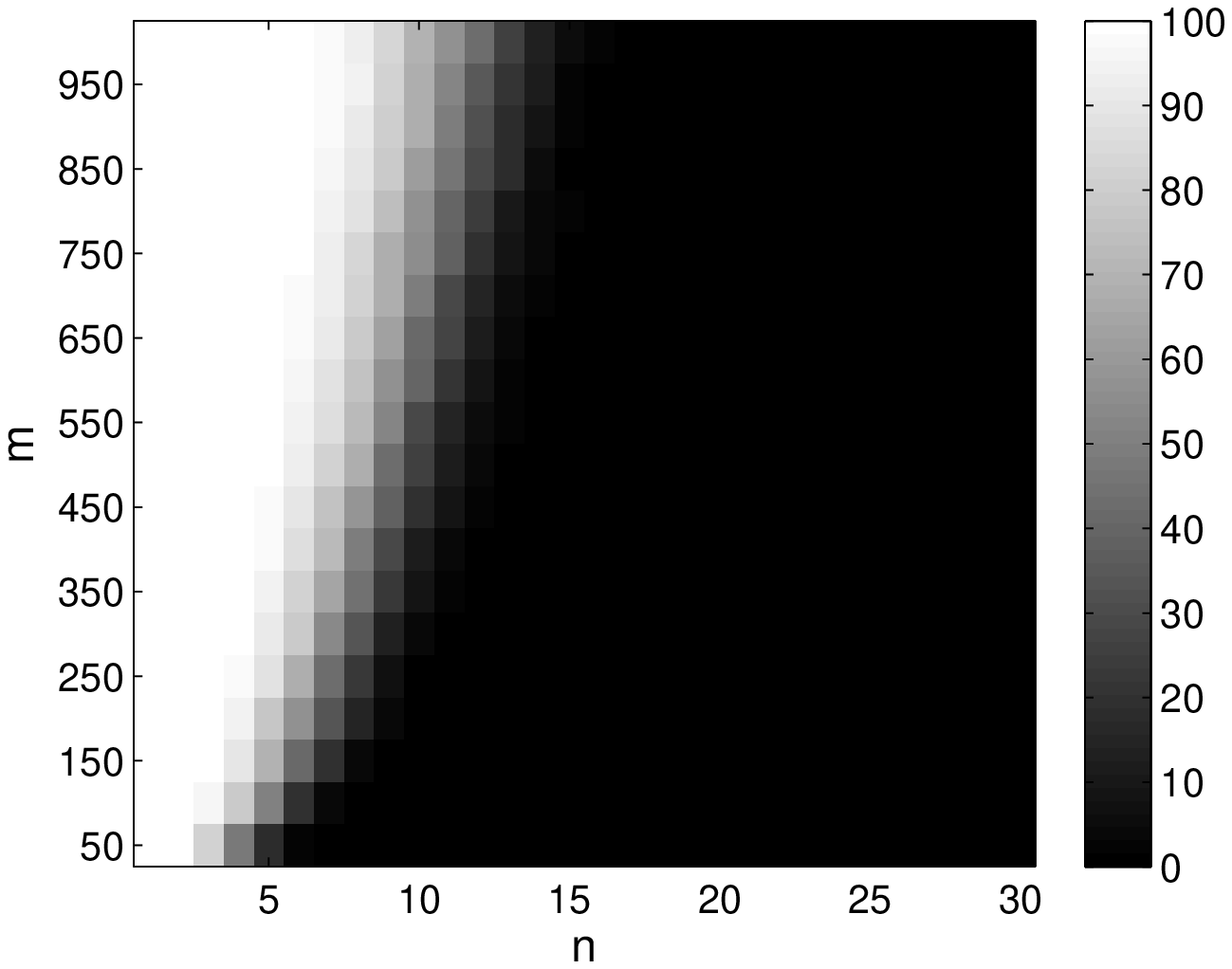}\\
normal&uniform\\
\includegraphics[width=7cm]{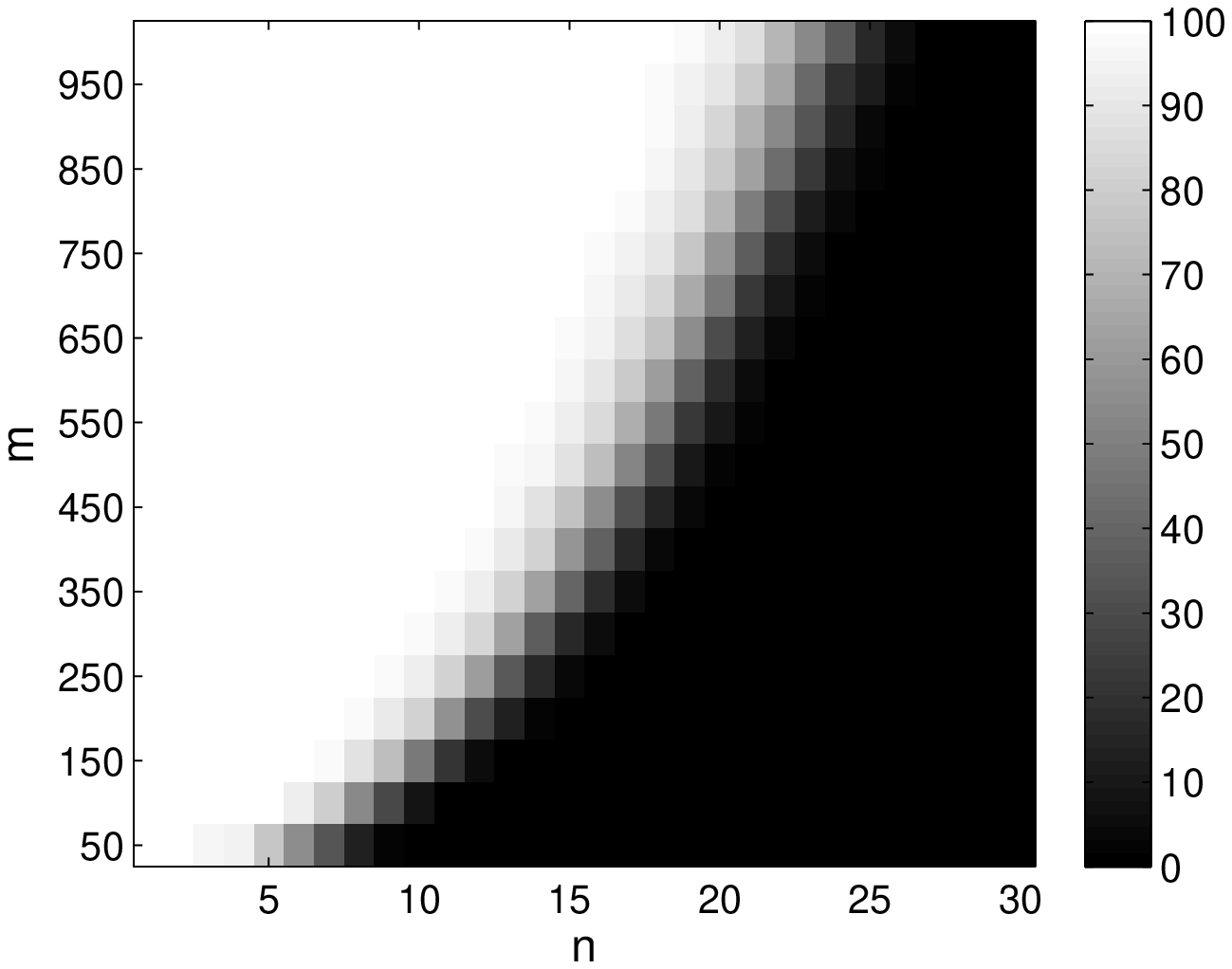}&\includegraphics[width=7cm]{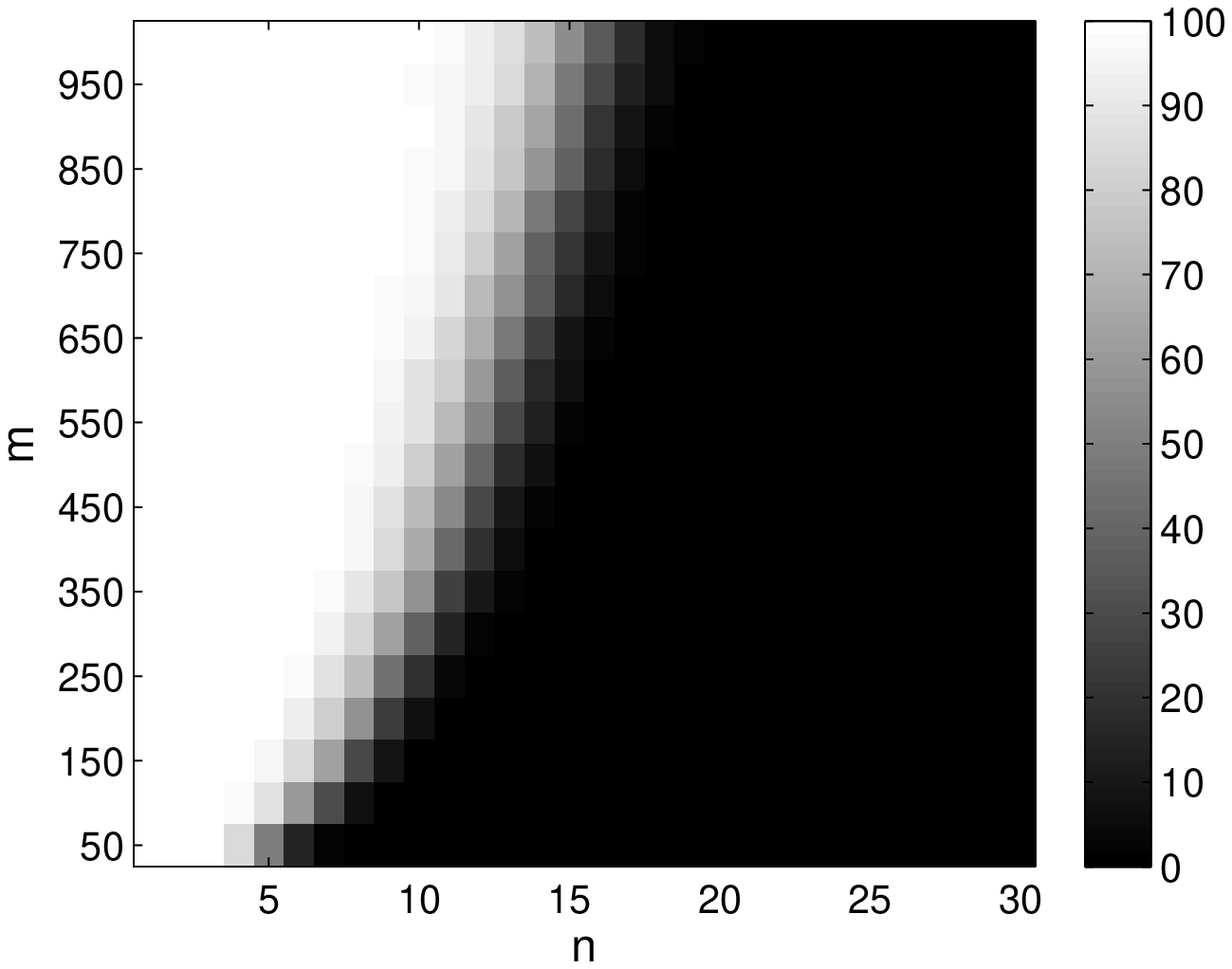}\\
Bernoulli with $p=0.1$&Bernoulli with $p=0.5$
\etabu
\caption{The frequency (in percentage) of $(\Ab^T\Ab)^{-1}$ being DD. Distribution \Pc is chosen as: normal distribution, uniform distribution within interval $[0,1]$ and Bernoulli distribution with parameter $p=0.1,0.5$.}\label{fig1}
\efig

%\btabl[t]\caption{The frequency (in percentage) of $(\Ab^T\Ab)^{-1}$ being IDD. Distribution \Pc is chosen as: normal distribution, uniform distribution within interval $[0,1]$ and Bernoulli distribution with parameter $p=0.1,0.3,0.5$.}\label{tab1}\centering
%\btabu{|c|c|c|c|c|c|}
%\hline
%$(m, n)\backslash~\Pc$&normal&uniform&Bernoulli~($p=0.1$)&Bernoulli~($p=0.3$)&Bernoulli~($p=0.5$)\\\hline
%(100,4)fig2&100&99.97&100&100&99.98\\\hline
%(100,6)&99.88&78.93&99.62&96.94&88.75\\\hline
%(100,8)&93.51&12.83&87.09&52.04&25.15\\\hline
%(100,10)&49.78&0.11&32.75&3.28&0.48\\\hline
%(100,12)&5.6&0&1.68&0&0\\\hline
%(1000,10)&100&99.81&100&100&99.98\\\hline
%(1000,20)&100&0.09&99.24&34.59&2.47\\\hline
%(1000,30)&76.61&0&0.02&0&0\\\hline
%(1000,40)&0&0&0&0&0\\\hline
%\etabu
%\etabl

\section{Connection with other conditions}\label{sec3}
\subsection{Connection with Donoho \etal's condition}\label{subsec1}
\paragraph*{$k$-step solution property}
For a given problem instance $(\Ab, \tilde{\yb})$, where $\Ab=[\ab_1,\cdots,\ab_n]\in\mathbb{R}^{m\times n}$, $\tilde{\yb}=\Ab\tilde{\ub}$, and $\tilde{\ub}$ has only $k$ nonzero entries. We say that an algorithm has $k$-step solution property at this given problem instance if it terminates after at most $k$-steps with the correct solution $\tilde{\ub}$.

In \cite{Donoho08}, Donoho gave a condition such that Homotopy algorithm has $k$-step solution property.

\paragraph*{Donoho \etal 's condition}
For the problem instance $(\Ab, \tilde{\yb})$, if the sparsity level $k$ obeys
\beq
k\leqslant\frac{1+\mu^{-1}}{2}
\eeq
where $\mu$ is the mutual coherence of \Ab:
\beq\nonumber
\mu=\max_{i\neq j}|<\ab_i,\ab_j>|
\eeq
then the Homotopy algorithm runs $k$ steps and stops, delivering the solution $\tilde{\ub}$. Here $<\cdot,\cdot>$ denotes the inner product.

In fact $\mu$ is the maximum of absolute value of off-diagonal entries of the Gram matrix $\Gb=\Ab^T\Ab$. Throughout this section, $\ab_i$ is normalized for convenience, \ie $\|\ab_i\|=1$. So the diagonal entry of \Gb is 1 and $\mu<1$.

As Homotopy algorithm was proved to be able to find the solution path of problem (\ref{homotopy}) \cite{Malioutov05}, Donoho \etal's condition can also be viewed as a sufficient condition which yields monotonic increase of $\Card{\Ic(\ub^*(\lambda))}$. However, Donoho \etal's condition need to know $k$, \ie the sparsity level of \ub, which is usually unknown in practical applications, while in Theorem \ref{thm1}, the knowledge of \ub is not needed.

Donoho \etal's condition reflects the following fact: lower correlated matrix \Ab (smaller $\mu$) yields more nonzero entries in \ub (larger $k$) that could be recovered. A natural deduction is for the limit case where $k=n-1$, which means \ub is not sparse at all; the upper bound of $\mu$ is $\frac{1}{2(n-1)-1}=\frac{1}{2n-3}(i\neq j)$, which is coincident with Corollary \ref{cor2} shown below.

\begin{theorem}\label{thm3}
Full rank symmetric matrix $\Gb \in \mathbb{R}^{n\times n}~(n>2)$, if $g_{ii}>0$ and $\frac{|g_{ij}|}{g_{ii}}\leqslant\frac{1}{2n-3}$, $\Gb^{-1}$ is DD.
\end{theorem}

\begin{proof}
If $\frac{|g_{ij}|}{g_{ii}}\leqslant\frac{1}{2n-3}~(j\neq i)$, $\sum_{j\neq i}\frac{|g_{ij}|}{g_{ii}}\leqslant\frac{n-1}{2n-3}<1$ for $n>2$ $\Rightarrow$ $\Gb$ is SDD $\Rightarrow$ \Gb is positive definite and nonsingular \cite{Bernstein05} $\Rightarrow$ its inverse $\Hb=\Gb^{-1}$ is also positive definte $\Rightarrow$ $h_{ii}>0$. From $\Hb\Gb=\Ib$ we have:
\beq\nonumber
\delta_{ij}=\sum_vh_{iv}g_{vj}=\sum_{v\neq j}h_{iv}g_{vj}+h_{ij}g_{jj}
\eeq
here $\delta_{ij}$ is kronecker symbol.
\beqn
\nonumber\sum_{j\neq i}|h_{ij}|&=&\sum_{j\neq i}\bars{\frac{\delta_{ij}-\sum_{v\neq j}h_{iv}g_{vj}}{g_{jj}}}\\
\nonumber&=&\sum_{j\neq i}\bars{-\sum_{v\neq j}\frac{g_{vj}}{g_{jj}}h_{iv}}\\
\nonumber&\leqslant&\frac{1}{2n-3}\sum_{j\neq i}\sum_{v\neq j}|h_{iv}|\\
\nonumber&=&\frac{1}{2n-3}\sum_{j\neq i}\left(\sum_v|h_{iv}|-|h_{ij}|\right)\\
\nonumber&=&\frac{1}{2n-3}\left(\sum_{j\neq i}\sum_v|h_{iv}|-\sum_{j\neq i}|h_{ij}|\right)\\
\nonumber&=&\frac{1}{2n-3}\left((n-1)\sum_v|h_{iv}|-\sum_{j\neq i}|h_{ij}|\right)\\
\nonumber&=&\frac{1}{2n-3}\left((n-1)\sum_{v\neq i}|h_{iv}|+(n-1)h_{ii}-\sum_{j\neq i}|h_{ij}|\right)\\
\nonumber&=&\frac{1}{2n-3}\left((n-2)\sum_{j\neq i}|h_{ij}|+(n-1)h_{ii}\right)
\eeqn
by moving $\sum_{j\neq i}|h_{ij}|$ in the right hand side to the left hand side, we have $\sum_{j\neq i}|h_{ij}|<|h_{ii}|=h_{ii}$, so $\Gb^{-1}=\Hb$ is DD.
\end{proof}

\begin{corollary}\label{cor2}
For symmetric matrix $\Gb \in \mathbb{R}^{n\times n}$ with $g_{ii}=1$,$|g_{ij}|\leqslant\frac{1}{2n-3}~(i\neq j)$, $\Gb^{-1}$ is DD.
\end{corollary}

\begin{remark}
As $\mu$ is equal to the maximum of absolute value of off-diagonal entries of the Gram matrix $\Gb=\Ab^T\Ab$. $\mu\leq\frac{1}{2n-3}$ yields $(\Ab^T\Ab)^{-1}$ being DD. Therefore, Donoho \etal's condition and Theorem \ref{thm1} are connected \via Corollary \ref{cor2}.
\end{remark}

\subsection{Connection with Efron \etal's positive cone condition}
\paragraph*{Positive cone condition} For each principal minor of $\Bb^T\Ab^T\Ab\Bb$, the sum of each row of the inverse matrix of this principal minor is positive. Here \Bb is the diagonal matrix whose diagonal entry is $\pm1$.

In \cite{Meinshausen06}, Meinshausen pointed out that Efron \etal's positive cone condition \cite{Efron04} yields monotonic increase of the absolute value of the LASSO estimator. In other words, the monotonic increase of the number of nonzero entry. In fact, from Lemma \ref{lemma1} we can deduce that the positive cone condition is equivalent to the condition that $(\Ab^T\Ab)^{-1}$ is SDD.

\begin{theorem}\label{thm4}
Positive cone condition is equivalent to the SDD condition on $(\Ab^T\Ab)^{-1}$.
\end{theorem}
\begin{proof}
\bit
\item Positive cone condition $\Rightarrow$ SDD condition on $(\Ab^T\Ab)^{-1}$

Each principal minor of $\Bb^T\Ab^T\Ab\Bb$ can be written as $\Jb_{k\times n}\Pb\Bb^T\Ab^T\Ab\Bb\Pb^T\Jb^T_{k\times n}$, positive cone condition demands that for any \Pb, \Bb and for all $k=1,\ldots,n$, the sum of each row of its inverse matrix should be positive. For the configuration where \Pb is the identity matrix and $k=n$, the sum of the $i$-th row of $(\Bb^T\Ab^T\Ab\Bb)^{-1}$, or $\Bb^T\Hb\Bb$ can be written as $\sum_{j=1}^nb_{ii}b_{jj}h_{ij}=h_{ii}+\sum_{j\neq i}b_{ii}b_{jj}h_{ij}$; the positive cone condition reads $h_{ii}+\sum_{j\neq i}b_{ii}b_{jj}h_{ij}>0$. Because $b_{ii}$ and $b_{jj}$ could be either $+1$ or $-1$, proper choice of $b_{ii}$ and $b_{jj}$ yields $h_{ii}>\sum_{j\neq i}|h_{ij}|$, \ie $\Hb=(\Ab^T\Ab)^{-1}$ is SDD.

\item SDD condition on $(\Ab^T\Ab)^{-1}$ $\Rightarrow$ positive cone condition

$\Hb=(\Ab^T\Ab)^{-1}$ being SDD yields $h_{ii}>\sum_{j\neq i}|h_{ij}|\geqslant\sum_{j\neq i}b_{ii}b_{jj}h_{ij}$ for any configuration of \Bb. So the positive cone condition is true for $k=n$. From Lemma \ref{lemma1}, \ie the DD (or SDD) preservation property, the inverse matrix of each principal minor of $\Ab^T\Ab$ is also SDD. So the positive cone condition is true for $k<n$.
\eit
\end{proof}

\begin{remark}
Because SDD condition is stronger than DD condition, from Theorem \ref{thm1} and Theorem \ref{thm4}, we find that the positive cone condition can be relaxed in order to have the monotonic increase of number of nonzero entry. In other words, the \textbf{positive} in positive cone condition can be relaxed to \textbf{nonnegative}.
\end{remark}

In practical applications the positive cone condition is difficult to test because of the huge number of configurations of both the principal minor and \Bb. On the contrary, the condition in Theorem \ref{thm1} is more practicable.

\section{Sufficient condition for total variation denoising}\label{sec4}
In signal processing community, the following total variation case is often used such as in denoising \cite{Chambolle97}.
\begin{equation}\label{homotopy2}
\xb^*(\lambda)=\argmin_{\xb}\acc{\frac{1}{2}\|\yb-\xb\|^2+\lambda\|\Db\xb\|_1}
\end{equation}
where $\Db$ could be chosen as the first order derivative matrix of size $(n-1)\times n$:
\beq\label{Dmatrix}
\left[
\barr{ccccc}
1&-1&0&\cdots&0\\
0&1&-1&\ddots&\vdots\\
\vdots&\ddots&\ddots&\ddots&0\\
0&\cdots&0&1&-1
\earr
\right]
\eeq
In the following, we present a sufficient condition, where \Db is not necessarily the first derivative matrix.

\begin{lemma}\label{lemma2}
For full rank $\Db\in\mathbb{R}^{m\times n}(m\leqslant n)$, problem (\ref{homotopy2}) is equivalent to the following one
\begin{equation}\label{OP3}
\ub^*(\lambda)=\argmin_\ub\acc{\frac{1}{2}\|\zb-\Ab\ub\|^2+\lambda\|\ub\|_1}
\end{equation}
where
\beq\label{def_uza}
\begin{tabular}{|rcl|}
  \hline
  $\ub$&=&$\Db\xb$\\
  $\zb$&=&$\Db^T(\Db\Db^T)^{-1}\Db\yb$\\
  $\Ab$&=&$\Db^T(\Db\Db^T)^{-1}$\\
  \hline
\end{tabular}
\eeq
\end{lemma}

\begin{proof}
The optimization problem (\ref{homotopy2}) is equivalent to the following constrained optimization problem
\begin{equation}\label{OP4}
%\nonumber (\xb^*(\lambda),\ub^*(\lambda))=\argmin_{\xb,\ub\atop s.t.\Db\xb=\ub}\acc{\frac{1}{2}\|\yb-\xb\|^2+\lambda\|\ub\|_1}
(\xb^*(\lambda),\ub^*(\lambda))=\argmin_{\xb,\ub}\acc{\frac{1}{2}\|\yb-\xb\|^2+\lambda\|\ub\|_1} \hspace{0.5cm}\hbox{subject to~~}\Db\xb=\ub
\end{equation}
The Lagrange function associated with (\ref{OP4}) reads
\begin{equation*}
L(\xb,\ub,\mub)=\frac{1}{2}\|\yb-\xb\|^2+\lambda\|\ub\|_1+\mub^T(\ub-\Db\xb)\\
\end{equation*}
where $\mub$ is Lagrange multiplier. The optimality condition reaches
\begin{eqnarray}
\nonumber\frac{\partial L}{\partial \xb}&=&\xb-\yb-\Db^T\mub=\zerob\\
\nonumber\frac{\partial L}{\partial \mub}&=&\ub-\Db\xb=\zerob
\end{eqnarray}
From above two equations, we have
\begin{eqnarray}
\nonumber\mub&=&(\Db\Db^T)^{-1}(\ub-\Db\yb)\\
\xb&=&\yb+\Db^T(\Db\Db^T)^{-1}(\ub-\Db\yb)\label{eqx}
\end{eqnarray}
by substituting (\ref{eqx}) into (\ref{OP4}), (\ref{OP4}) rereads
\beq
\nonumber\ub^*(\lambda)=\argmin_\ub\acc{\frac{1}{2}\|\Db^T(\Db\Db^T)^{-1}(\Db\yb-\ub)\|^2+\lambda\|\ub\|_1}
\eeq
which is the same as (\ref{OP3}) where \ub, \zb, and \Ab are defined as in (\ref{def_uza}). So (\ref{homotopy2}) is equivalent to (\ref{OP3}).
\end{proof}

\begin{theorem}\label{thm2}
For full rank matrix $\Db\in\mathbb{R}^{m\times n}(m\leqslant n)$ and optimization problem (\ref{homotopy2}), if $\Db\Db^T$ is DD, $\Card{\Ic(\xb^*(\lambda))}$ \textbf{increases} monotonically when $\lambda$ \textbf{decreases}.
\end{theorem}

\begin{proof}
From Lemma 2, $(\Ab^T\Ab)^{-1}=\Db\Db^T$ is DD. By applying Theorem \ref{thm1}, we get this theorem straightforwards.
\end{proof}

It is easy to verify that the first derivative matrix (\ref{Dmatrix}) satisfies the condition in Theorem \ref{thm2}. Thus, the results hold for the optimization problem (\ref{homotopy2}) with total variation case.

\section{Conclusion\label{Conclu}}
In this paper, we presented a sufficient condition under which the number of nonzero entries in the optimizer of $\ell$-1 norm penalized least-square problem increases monotonically. Sufficient condition for the total variation case is also presented. We showed that the sufficient condition, \ie the inverse of the Gram matrix of the matrix \Ab is diagonally dominant, is strongly connected with Donoho \etal's condition and is equivalent to or more general than Efron \etal's positive cone condition. Compared with Donoho \etal's condition which yields $k$-step solution property, our proposed condition does not need the knowledge of the original signal (\ie the sparsity level), which is usually unknown in practical application. Compared with Efron \etal's positive cone condition which needs to test a large number of configurations in an exhaustive manner, our proposed condition is simpler to verify.

\appendix
\section{Proof of Lemma \ref{lemma1}}\label{apx1}
In order to prove Lemma \ref{lemma1}, we introduce the following two Lemmas.
\begin{lemma}\label{lemma3}
If full rank symmetric matrix $\Hb \in \mathbb{R}^{n\times n}$ is DD, then $\Rb=(\Jb_{(n-1)\times n}\Hb^{-1}\Jb_{(n-1)\times n}^T)^{-1}$ is also DD.
\end{lemma}

\begin{proof}
Define:
\begin{equation*}
    \Hb= \left[\begin
    {array}{cc} \Hb_{11}&\Hb_{12}\\\Hb_{12}^T&\Hb_{22}
    \end {array}\right]
\end{equation*}
\begin{equation*}
    \Gb=\Hb^{-1}= \left[\begin
    {array}{cc} \Gb_{11}&\Gb_{12}\\\Gb_{12}^T&\Gb_{22}
    \end {array}\right]
\end{equation*}
where $\Hb_{11}$, $\Gb_{11}$ are of size $(n-1)\times(n-1)$. The sub-matrices can be expressed to $\Hb$ and $\Gb$ according to
\begin{eqnarray}\label{eq_HG}
\nonumber\Hb_{11}&=&\Jb_{(n-1)\times n}\Hb\Jb^T_{(n-1)\times n}\\
\nonumber\Hb_{12}&=&[h_{1n},h_{2n}\cdots h_{n-1,n}]^T\\
\Hb_{22}&=&[h_{nn}]\\
\nonumber\Gb_{11}&=&\Jb_{(n-1)\times n}\Gb\Jb^T_{(n-1)\times n}\\
\nonumber\Gb_{12}&=&[g_{1n},g_{2n}\cdots g_{n-1,n}]^T\\
\nonumber\Gb_{22}&=&[g_{nn}]
\end{eqnarray}
From block matrix inversion lemma \cite{Bernstein05}, we know
\beq\nonumber
\Rb=\Gb^{-1}_{11}=\Hb_{11}-\Hb_{12}\Hb_{22}^{-1}\Hb_{12}^T
\eeq
From above and (\ref{eq_HG}), the entry of $\Rb$ reads
\beq
\nonumber r_{ij}=h_{ij}-\frac{h_{in}h_{jn}}{h_{nn}},(i=1\ldots n-1,j=1\ldots n-1)
\eeq
then we have
\begin{eqnarray}
\sum_{j\neq i,n}|r_{ij}|
\nonumber&=&\sum_{j\neq i,n}\left|h_{ij}-\frac{h_{in}h_{jn}}{h_{nn}}\right|\\
\nonumber&\leqslant&\sum_{j\neq i,n}|h_{ij}|+\frac{|h_{in}|}{h_{nn}}\sum_{j\neq i,n}|h_{jn}|\\
\nonumber&\stackrel{\mathrm{DD}}{\leqslant}&\sum_{j\neq i,n}|h_{ij}|+\frac{|h_{in}|}{h_{nn}}(h_{nn}-|h_{in}|)\\
\nonumber&=&\sum_{j\neq i,n}|h_{ij}|+|h_{in}|-\frac{h_{in}^2}{h_{nn}}\\
\nonumber&=&\sum_{j\neq i}|h_{ij}|-\frac{h_{in}^2}{h_{nn}}\\
\nonumber&\stackrel{\mathrm{DD}}{\leqslant}&h_{ii}-\frac{h^2_{in}}{h_{nn}}\\
\nonumber&=&r_{ii}
\end{eqnarray}
So $\Rb$ is DD.
\end{proof}

\begin{lemma}\label{lemma4}
If full rank symmetric matrix $\Hb \in \mathbb{R}^{n\times n}$ is DD, then $\Rb_k=(\Jb_{k\times n}\Hb^{-1}\Jb_{k\times n}^T)^{-1}$ is also DD for all $k=1,\ldots,n-1$.
\end{lemma}

\begin{proof}
$\Hb$ is full rank symmetric, so $\Rb_{i},(i=1\ldots n-1)$ is also full rank symmetric. By using Lemma \ref{lemma3} recursively:

$\Hb$ is DD $\Rightarrow$ $\Rb_{n-1}$ is DD $\Rightarrow$ $\Rb_{n-2}$ is DD $\Rightarrow\cdots\cdots\Rightarrow$ $\Rb_1$ is DD.
\end{proof}

\begin{proof}[\textbf{Proof of Lemma \ref{lemma1}}]
\Hb is full rank symetric DD, so $\Pb\Hb\Pb^T$ is also full rank symetric DD. From Lemma \ref{lemma4}, Lemme \ref{lemma1} is straightforward.
\end{proof}

\bibliography{gpipubli,revuedef,revueabr,baseAJ,baseKZ,ref_db,bibenabr}
\end{document}